\documentclass[10pt,twocolumn,letterpaper]{article}

\usepackage[pagenumbers]{iccv} 

\usepackage{cuted}
\usepackage[accsupp]{axessibility}  

\definecolor{iccvblue}{rgb}{0.21,0.49,0.74}
\usepackage[pagebackref,breaklinks,colorlinks,allcolors=iccvblue]{hyperref}

\title{ID-Consistent, Precise Expression Generation with Blendshape-Guided Diffusion}

\author{Foivos Paraperas Papantoniou
\quad \quad \quad
Stefanos Zafeiriou\\[6pt]
Imperial College London, UK
\\{\tt\small \{f.paraperas, s.zafeiriou\}@imperial.ac.uk}
}

\begin{document}
\maketitle

\begin{strip}
\centering
\includegraphics[width=\textwidth]{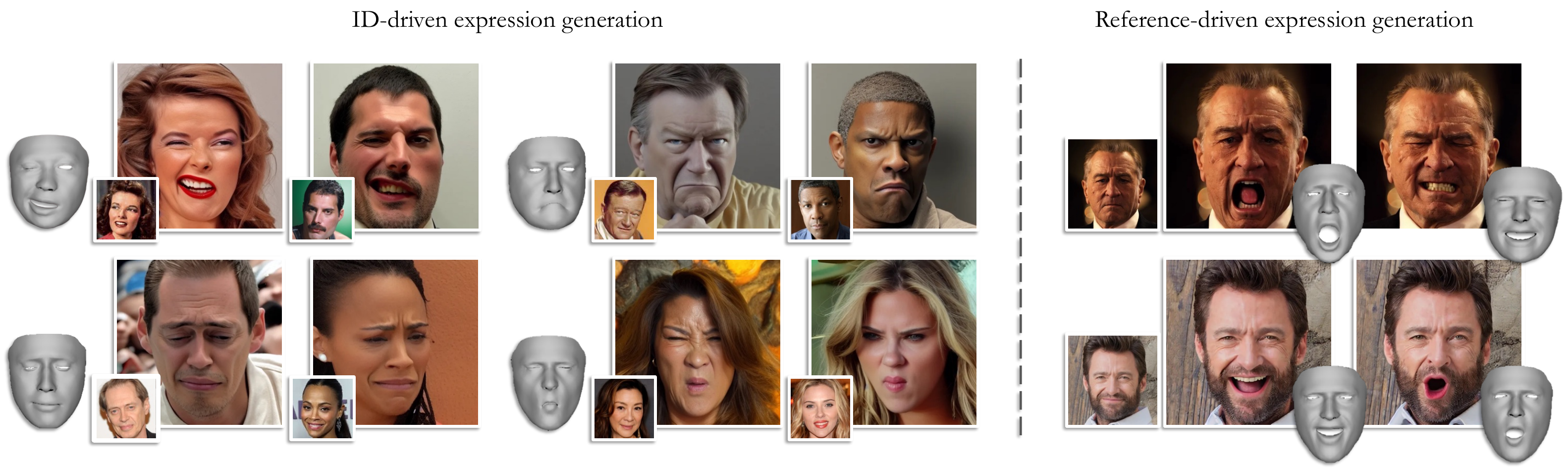}
\captionof{figure}{We introduce a fine-grained expression adapter on a foundation ID-consistent face model, which significantly outperforms existing approaches in expression-transfer fidelity and can apply any type of facial expression to any given subject - including extreme or asymmetric ones - using explicit blendshape parameters (left). Our method can optionally integrate a reference adapter that enables expression editing without altering the appearance or background (right).}
\label{fig:teaser}
\end{strip}

\begin{abstract}
Human-centric generative models designed for AI-driven storytelling must bring together two core capabilities: identity consistency and precise control over human performance. While recent diffusion-based approaches have made significant progress in maintaining facial identity, achieving fine-grained expression control without compromising identity remains challenging. In this work, we present a diffusion-based framework that faithfully reimagines any subject under any particular facial expression. Building on an ID-consistent face foundation model, we adopt a compositional design featuring an expression cross-attention module guided by FLAME blendshape parameters for explicit control. Trained on a diverse mixture of image and video data rich in expressive variation, our adapter generalizes beyond basic emotions to subtle micro-expressions and expressive transitions, overlooked by prior works. In addition, a pluggable Reference Adapter enables expression editing in real images by transferring the appearance from a reference frame during synthesis. Extensive quantitative and qualitative evaluations show that our model outperforms existing methods in tailored and identity-consistent expression generation. Code and models can be found at \url{https://github.com/foivospar/Arc2Face}.
\end{abstract}    
\section{Introduction}
\label{sec:intro}

The rapid advancement of generative models, beginning with GANs \cite{goodfellow2020generative} and now dominated by diffusion models \cite{ddpm_Ho, song2021Score-SDE, dhariwal2021diffusion}, has significantly elevated the quality and versatility of synthetic visual content. Powered by high-performance computing and trained on vast amounts of data, foundation diffusion models can generate high-quality and diverse images of human characters, achieving remarkable realism and adaptability across domains. The impact of such technology spans multiple industries, from advertising to education and virtual environments. Yet, one of the most transformative applications lies in film and entertainment, where human performance is central to storytelling. Despite progress in human-centric generation, the existing paradigm in diffusion models focuses on identity preservation, often overlooking controllability.

Facial expressions, however, play a vital role in human communication, conveying emotions, intentions, and personality traits. As such, they have become a central focus in face modeling research. A long line of work has explored facial expression editing with GAN-based methods \cite{StarGAN, sdapolito2021GANmut, ding2017exprgan, Tripathy_ICface, paraperas2022ned}, though often with limited realism. More recent approaches leverage powerful pre-trained text-to-image diffusion models like Stable Diffusion \cite{rombach2022high}, extending them with additional mechanisms for identity and expression control to generate faces with desired emotional states. These methods typically rely on emotion representations such as categorical labels (\eg, happy, sad), the continuous Valence-Arousal 2D space \cite{russell1977evidence}, or Action Units (AUs) \cite{ekman1978facial} that describe localized facial muscle movements. However, while these descriptors offer semantic editing, they often fall short in capturing the full precision and nuance of facial expressions, limiting their usefulness for applications requiring fine-grained control.

In this work, we target precise and disentangled facial expression control. To achieve this, we adopt a parametric representation based on expression blendshapes from the FLAME 3D face model \cite{FLAME:SiggraphAsia2017}, which offers continuous, high-dimensional control over expression. We build upon Arc2Face \cite{paraperas2024arc2face}, a foundation diffusion model which can synthesize diverse, realistic faces of a single identity using identity embeddings as conditioning input. To incorporate detailed control, we inject the expression parameters into the model's cross-attention layers, while the pretrained backbone remains frozen, preserving its strong identity prior. To support a wide range of expressive capabilities, including rare, asymmetric, subtle, and extreme expressions, we carefully curate training datasets with rich expression diversity. Additionally, we introduce a Reference Adapter trained on top of the base model that conditions generation on an input image, allowing precise expression editing while maintaining the subject’s original appearance and background.

\section{Related Work}
\label{sec:related_work}

\subsection{Facial Expression Generation and Editing}

Early efforts to edit facial expressions in ``in-the-wild'' images primarily employed GAN-based architectures. These approaches typically used conditional GANs for image-to-image translation. For instance, StarGAN \cite{StarGAN} enabled multi-domain facial attribute transfer using categorical emotion labels, while ExprGAN \cite{ding2017exprgan} and Lindt \etal \cite{Lindt2019FacialEE} introduced expression editing along emotion intensity levels. Deviating from hand-crafted emotion representations, GANmut \cite{sdapolito2021GANmut} proposed a framework that implicitly learns a continuous and interpretable emotion space from categorical labels, by associating classification confidence with expression intensity. Despite promising results, these methods are limited by the quality and diversity of their training data, as well as the inherent instability of GAN training. 

At the same time, various works focus on geometry-driven expression editing, addressing the task of face reenactment. Models like ICface \cite{Tripathy_ICface} and GANimation \cite{GANimation_ijcv2019} leverage Action Units (AUs) as an interpretable representation for expression control, whereas Neural Emotion Director \cite{paraperas2022ned} introduces a 3D-based Emotion Manipulator that translates expressions into basic emotions or reference-driven styles using a 3D Morphable Model (3DMM)-conditioned GAN. Another direction explores the latent space of the pre-trained StyleGAN \cite{stylegan} network, for facial editing. InterFaceGAN \cite{shen2020interpreting, shen2020interfacegan} disentangles facial attributes using pre-trained classifiers, enabling semantic control over identity, age, or expression. StyleCLIP \cite{patashnik2021styleclip} builds on this by incorporating CLIP \cite{radford2021learning}, allowing text-driven facial edits within StyleGAN’s latent space, while EmoStyle \cite{EmoStyle_2024_WACV} takes this further by explicitly separating expression from other facial attributes, using Valence-Arousal values to modify expressions while preserving identity and appearance.

\subsection{ID and Attribute Control in Diffusion Models}

Following the advent of diffusion models, research has shifted towards adding controllability to frontier text-to-image models like Stable Diffusion \cite{rombach2022high}. One major area of focus has been identity preservation. A popular approach involves using the CLIP image encoder \cite{radford2021learning} to extract features from reference subjects, which are then injected into pre-trained models via cross-attention layers. Notable examples include FastComposer \cite{xiao2023fastcomposer}, PhotoVerse \cite{chen2023photoverse}, MoA \cite{wang2024moa}, and PhotoMaker \cite{li2023photomaker}. A more robust alternative to CLIP embeddings is the use of face recognition networks, which extract stronger ID-specific features. This approach has been leveraged in works such as Face0 \cite{valevski2023face0}, DreamIdentity \cite{chen2023dreamidentity}, IP-Adapter-FaceID \cite{ye2023ip}, PortraitBooth \cite{peng2023portraitbooth}, and InstantID \cite{wang2024instantid} for both 2D and 3D face generation \cite{gerogiannis2025arc2avatar}. Among these, the pioneering Arc2Face work~\cite{paraperas2024arc2face}, showed how the powerful Stable Diffusion model can be transformed into an ID-consistent face foundation model capable of generating highly realistic and diverse images with compelling identity similarity, by leveraging WebFace \cite{zhu2021webface260m} - the largest public dataset for face recognition. 

In parallel, numerous efforts explore the integration of expression features. These methods vary in the type of representation used, including 2D landmarks \cite{han2024faceadapter, mishima2025facecrafter}, 3DMMs \cite{liang2024caphuman}, Action Units \cite{wei2025magicface, varanka2024fineface}, continuous emotion spaces \cite{emotiondiffusion}, or natural language instructions \cite{wang2024instructavatar, liu2024towards}. Yet, they often fall short in expression fidelity - particularly when handling asymmetric or extreme expressions and typically introduce identity distortion.
\section{Method}
\label{sec:method}

\begin{figure*}[h]
\centering
\includegraphics[width=1.0\textwidth]{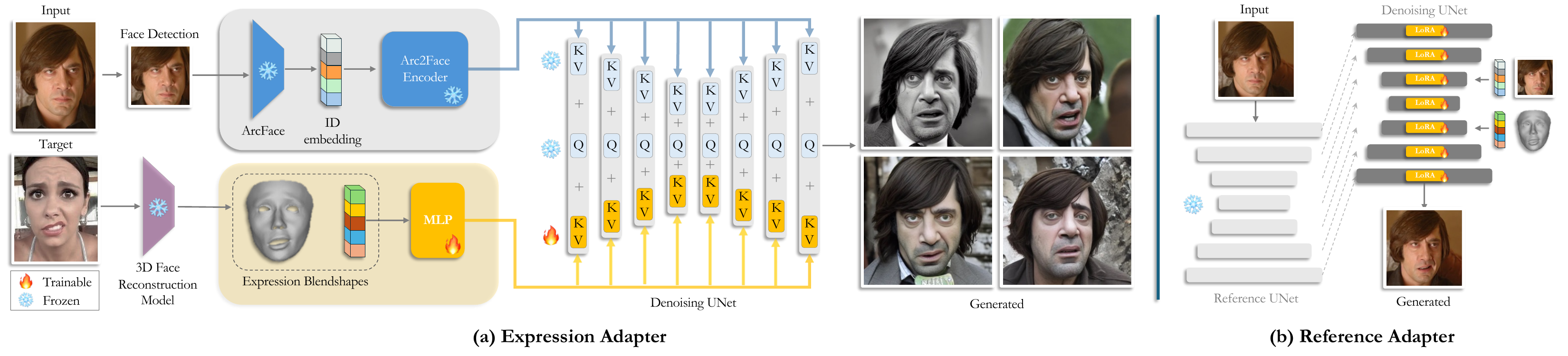}
\caption{\textbf{Overview of the proposed expression-control framework.} Our approach builds on the Arc2Face diffusion model~\cite{paraperas2024arc2face}, which conditions the denoising UNet on ID embeddings. \textbf{(a)} We introduce an \textbf{Expression Adapter} that guides the generation using explicit FLAME~\cite{FLAME:SiggraphAsia2017} blendshape parameters extracted from reference images via an off-the-shelf 3D reconstruction method~\cite{smirk_2024_CVPR}. The adapter consists of two components: (1) an MLP that maps 3DMM parameters into the CLIP latent space used by Stable Diffusion, and (2) a dual attention mechanism that integrates expression information alongside identity using separate key and value matrices in the cross-attention layers. \textbf{(b)} We further incorporate a \textbf{Reference Adapter} that conditions the model on the input image itself. A dedicated Reference UNet extracts multi-scale features, which are injected into the denoising UNet via self-attention layers modulated by added LoRA weights. By enabling this adapter at inference, we support image-based expression editing. The proposed modules are trained on expression-rich image and video data, achieving strong generalization across a wide range of facial expressions.}
\label{fig:method}
\end{figure*}

Our method generates facial images with disentangled control over identity and expression. Given an input facial identity, represented by ArcFace \cite{deng2019arcface} embeddings, and a target expression, encoded as FLAME blendshape parameters extracted from an expressive target image, our diffusion model synthesizes diverse and photorealistic images that faithfully reflect the source subject with the target expression, as shown in \cref{fig:teaser}. To further enhance applicability, we introduce an optional inference-time mechanism that conditions the generation directly on the source image pixels, enabling expression editing on real images while preserving background and appearance. Specific details of our approach are provided in the subsequent sections. For a high-level overview, see \cref{fig:method}.

\subsection{Preliminary: Arc2Face}
Arc2Face \cite{paraperas2024arc2face} is a recent diffusion foundation model for human faces, built on top of the Stable Diffusion framework. It achieves identity-consistent synthesis by integrating ArcFace \cite{deng2019arcface} embeddings into the generation pipeline, enabling the creation of diverse, high-quality $512 \times 512$ facial images with a significantly higher degree of identity similarity compared to prior methods.  Starting from a face image $\mathbf{x}\in \mathbb{R}^{H\times W\times C}$ as input, the ArcFace network \cite{deng2019arcface} $\phi$ is used, following cropping and alignment, to extract the identity features, $\mathbf{v} = \phi(\mathbf{x}) \in \mathbb{R}^{512}$. In order to project these features into the cross-attention layers of the UNet for conditioning, Arc2Face repurposes the original CLIP text encoder $\tau$ to act as a facial identity encoder tailored to ArcFace embeddings. The identity vector $\mathbf{v}$ replaces a placeholder token within a fixed text prompt and is mapped by the fine-tuned encoder to a sequence in the CLIP latent space: $\mathbf{v}_c = \tau(\mathbf{v}) \in \mathbb{R}^{77 \times 768}$. The entire model - comprising the adapted CLIP encoder and the UNet - is extensively fine-tuned on WebFace260M \cite{zhu2021webface260m}, the largest publicly available face recognition dataset, enabling Arc2Face to function as a powerful prior model. Its strong decoupling between identity and other visual attributes makes it well-suited for our task, which aims to introduce fine-grained, expression-level control while maintaining subject fidelity.

\subsection{Expression Adapter}
To achieve fine-grained and disentangled control over facial expressions, we leverage a parametric 3DMM, specifically FLAME \cite{FLAME:SiggraphAsia2017}, which decomposes facial geometry into distinct components for identity and expression. This formulation maps the emotion control problem from image space to the subject-agnostic parameter space of a 3D model. We construct an expression representation vector by concatenating FLAME’s expression parameters $\mathbf{e} \in \mathbb{R}^{50}$, eyelid pose parameters $\mathbf{a} \in \mathbb{R}^{2}$, and jaw articulation parameters $\mathbf{p} \in \mathbb{R}^{3}$. The resulting vector $\mathbf{s} = \text{concat}(\mathbf{e}, \mathbf{a}, \mathbf{p}) \in \mathbb{R}^{55}$ serves as the input to our \textbf{Expression Adapter}.

Our approach follows the paradigm of IP-Adapter \cite{ye2023ip}, and integrates expression information into the cross-attention layers of the UNet in a disentangled manner. First, we use a lightweight MLP $\epsilon$ to project the expression vector into the CLIP latent space $\mathbf{s}_c = \epsilon(\mathbf{s}) \in \mathbb{R}^{4 \times 768}$. This projected embedding is then used to produce additional key and value matrices, $\mathbf{K}_{exp}, \mathbf{V}_{exp}$, which are incorporated into each cross-attention layer of the UNet. Specifically, at each layer, we extend the original cross-attention mechanism $\text{Attention}_{\text{id}}(\mathbf{Q}, \mathbf{K}, \mathbf{V})$ - conditioned on the identity embedding - with an additional branch conditioned on expression, where the two attention outputs are computed in parallel using the same query matrix $\mathbf{Q}$. The final attention output is the sum of both contributions:
\begin{equation}
\begin{split}
    \mathbf{S} = \text{Attention}_{\text{id}}(\mathbf{Q}, \mathbf{K}, \mathbf{V}) + \text{Attention}_{\text{exp}}(\mathbf{Q}, \mathbf{K}_{exp}, \mathbf{V}_{exp}) \\
    = \text{Softmax}(\frac{\mathbf{Q}\mathbf{K}^{\top}}{\sqrt{d}})\mathbf{V}+\text{Softmax}(\frac{\mathbf{Q}(\mathbf{K}_{exp})^{\top}}{\sqrt{d}})\mathbf{V}_{exp}
\end{split}
\end{equation}
To maintain the strong prior of the pre-trained Arc2Face backbone, we freeze all original model weights and train only the additional parameters: the expression projection MLP $\epsilon$, and the linear projection layers $\mathbf{W}_{k_{exp}}$, $\mathbf{W}_{v_{exp}}$ used to compute the expression key and value matrices $\mathbf{K}_{exp}=\mathbf{s}_{c}\mathbf{W}_{k_{exp}}, \mathbf{V}_{exp}=\mathbf{s}_{c}\mathbf{W}_{v_{exp}}$. 

\subsection{Reference Adapter}
While our base model enables stochastic portrait generation conditioned on identity and expression features, certain use cases, such as real image editing, require preserving the subject’s appearance and background. To support this, we introduce a \textbf{Reference Adapter}, which can be seamlessly integrated into the base model to condition the generation on the source image itself. This is implemented via a Reference UNet, a frozen copy of the original denoising UNet that mirrors its architecture. The Reference UNet serves as a feature encoder for the input reference image, extracting spatially-aligned representations that capture the subject’s appearance and background at each layer. Since these features are dimensionally aligned with those produced by the main (expression-conditioned) UNet during the denoising process, we concatenate the two feature maps within the self-attention layers. This allows the model to blend appearance cues from the reference image with the evolving features of the generated image. While this modification achieves reference conditioning at inference time, in practice, it may introduce conflicts with expression control - especially when the reference expression differs from the target. To resolve this, we incorporate lightweight LoRA layers into the self-attention modules of the denoising UNet. These are fine-tuned with both the \textbf{Expression Adapter} and the \textbf{Reference Adapter} active, enabling the model to harmonize identity, expression, and appearance without retraining the full network. During inference, our approach supports flexible generation: the model can switch between reference-driven and stochastic expression synthesis by adding or removing the Reference UNet and the optimized LoRA weights.

\subsection{Training Details}
Training is performed in two stages. In the first phase, we train the \textbf{Expression Adapter} using expression-rich datasets. We use AffectNet \cite{mollahosseini2017affectnet} with 450K images which we upsample with a face restoration network \cite{wang2021gfpgan}, and augment it with 250K frames from the FEED video dataset \cite{Drobyshev_2024_CVPR} containing extreme expressions. We also include 70K high-resolution images from FFHQ \cite{stylegan} for enhanced quality. For each image, we extract identity embeddings using ArcFace \cite{deng2019arcface} and expression parameters using the state-of-the-art SMIRK method \cite{smirk_2024_CVPR}, for conditioning, while the image itself is used as the supervision target for the denoising loss.

The second phase includes adding the LoRA layers of the \textbf{Reference Adapter}, and training them while keeping the trained \textbf{Expression Adapter} frozen. This requires cross-paired training samples where the same individual appears with different expressions in the source and target images, as using the same image would lead the model to simply ``copy-paste'' the reference image, bypassing the target expression. Therefore, we utilize the FEED video dataset, as well as the large-scale HDTF video dataset \cite{zhang2021hdtf}, containing approximately 1.5 million frames of talking faces. We sample reference–target pairs by selecting two random frames within short (2-second) clips, ensuring expression variation while maintaining identity consistency. Again, we annotate the reference images with identity embeddings and the target images with expression parameters.
\section{Experiments}
\label{sec:experiments}

\begin{figure*}[h]
\centering
\includegraphics[width=1.0\textwidth]{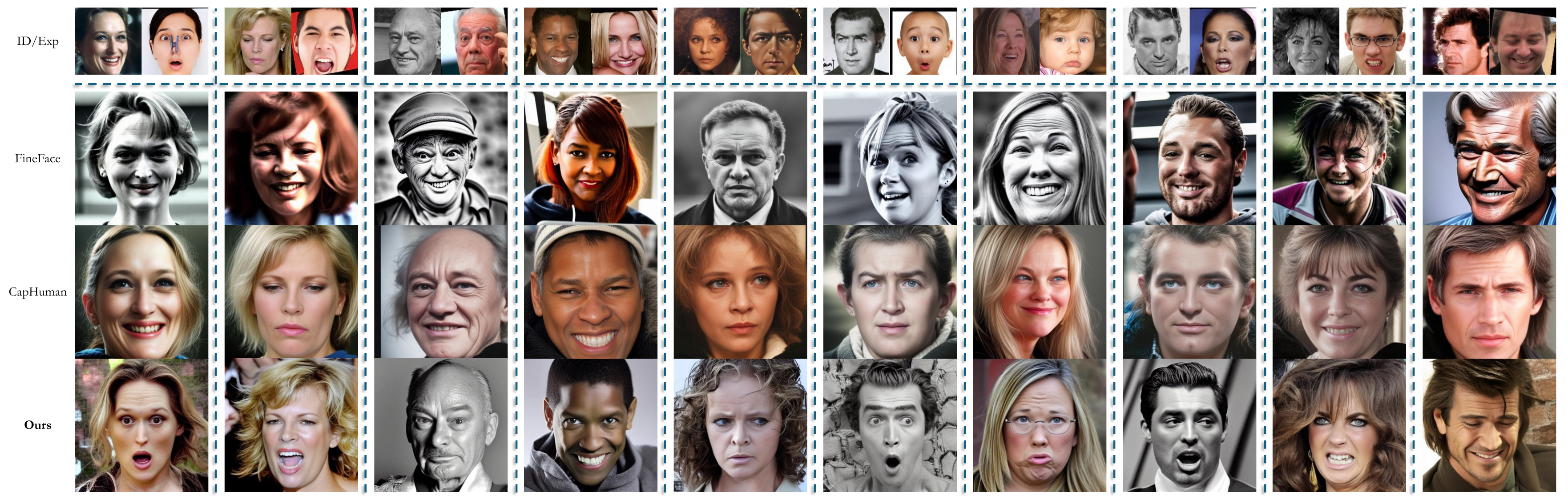}
\caption{Visual comparison of our method with competing models \cite{varanka2024fineface, liang2024caphuman} for expression-controlled generation conditioned on identity features.}
\label{fig:comp}
\end{figure*}

We conduct comprehensive comparisons with state-of-the-art open-source approaches for face generation with emotion control. Below, we outline the experimental setup, evaluation metrics, and baseline methods used in each setting. Additional qualitative results are provided in the Supplementary Material.

\subsection{Identity-Driven Expression Generation}
We first focus on the unconstrained generation setting, where the goal is to synthesize novel expressive facial images, conditioned on a given identity and target expression. We compare against recent similar methods that achieve expression control on top of Stable Diffusion: FineFace~\cite{varanka2024fineface}, which uses Action Units (AUs) as expression guidance, and CapHuman~\cite{liang2024caphuman}, which leverages rendered normals and albedo maps from FLAME with a ControlNet-like \cite{controlnet} structure. Both incorporate identity embeddings to preserve subject identity. For benchmarking, we collect 1K ``in-the-wild'' face images of different individuals. For each subject, we randomly sample 5 target expression images from the AffectNet validation set~\cite{mollahosseini2017affectnet}, resulting in 5K expression-conditioned generations per method. We evaluate performance across three main axes: image quality, expression fidelity, and identity preservation, using the following metrics: 1) \textbf{FID}~\cite{heusel2017gans, Seitzer2020FID}: Fréchet Inception Distance to assess realism by comparing distributions of generated and input identity images; 2) \textbf{VA-MSE}: MSE of Valence-Arousal values between generated and target expression images; 3) \textbf{AU-MSE}: MSE between target and generated Action Unit intensities; 4) \textbf{Emotion Accuracy}: Classification accuracy for 8 emotion categories (anger, contempt, disgust, fear, happiness, neutral, sadness, surprise) between intended and generated emotions; 5) \textbf{3D Exp-MSE}: MSE of reconstructed FLAME expression parameters using \cite{smirk_2024_CVPR} between target and synthesized images; 6) \textbf{ID Similarity}: Cosine similarity of identity embeddings between source and generated faces. To extract Valence-Arousal values, AU intensities, and emotion labels for all images, we employ off-the-shelf models from the Facetorch toolkit~\cite{facetorch, kim2022optimal, luo2022learning, Savchenko_2021}.

\begin{figure}[t]
  \begin{minipage}[t]{0.55\linewidth}
    \centering
    \includegraphics[width=\linewidth, trim={2.5cm 0.8cm 1.8cm 0.5cm}, clip]{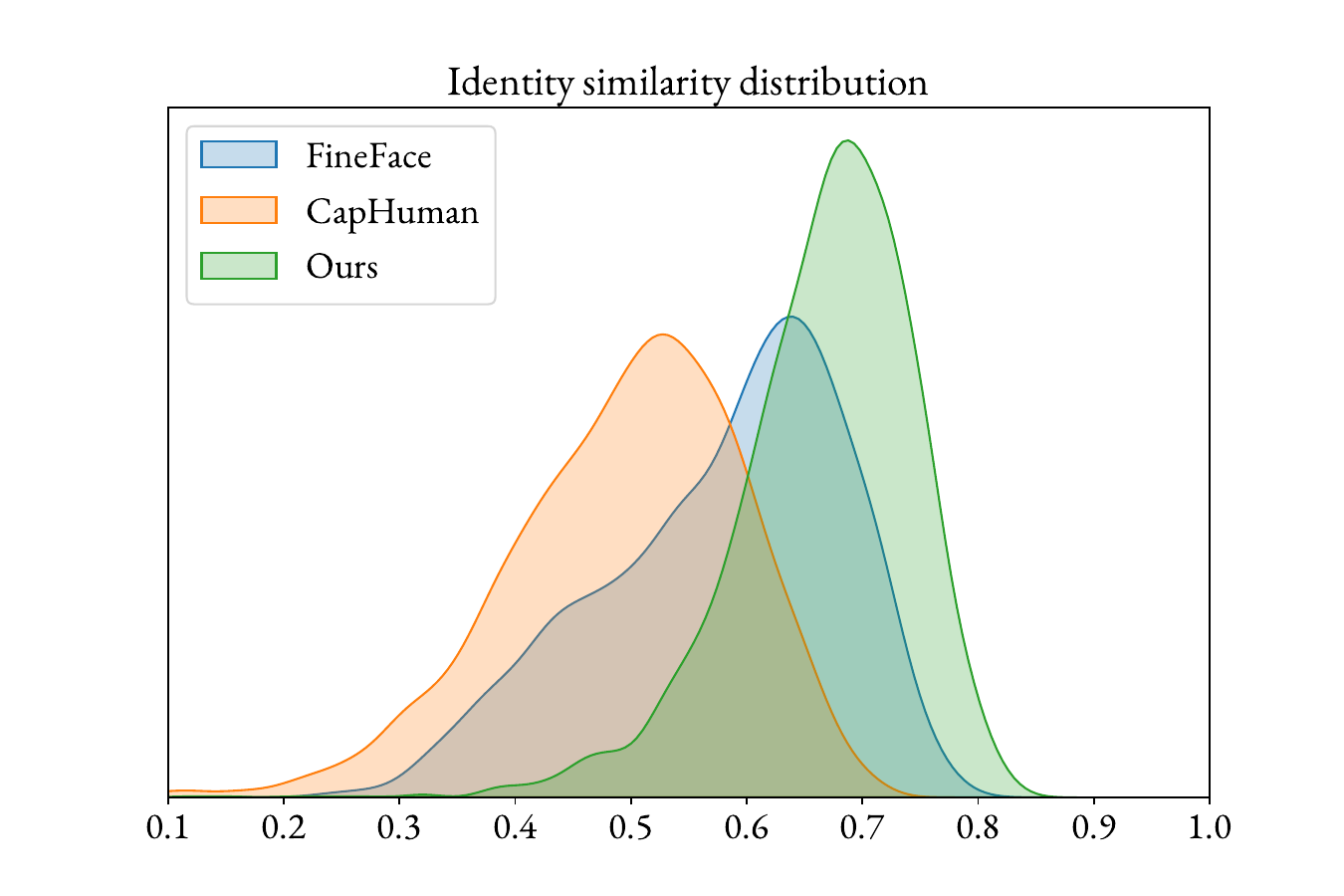}
    \caption{Cosine similarity distribution between identity features of input and generated faces.}
    \label{fig:id_sim}
  \end{minipage}%
  \hfill
  \begin{minipage}[t]{0.42\linewidth}
    \centering
    \includegraphics[width=0.88\linewidth, trim={2.5cm 1.0cm 1.0cm 0.5cm}, clip]{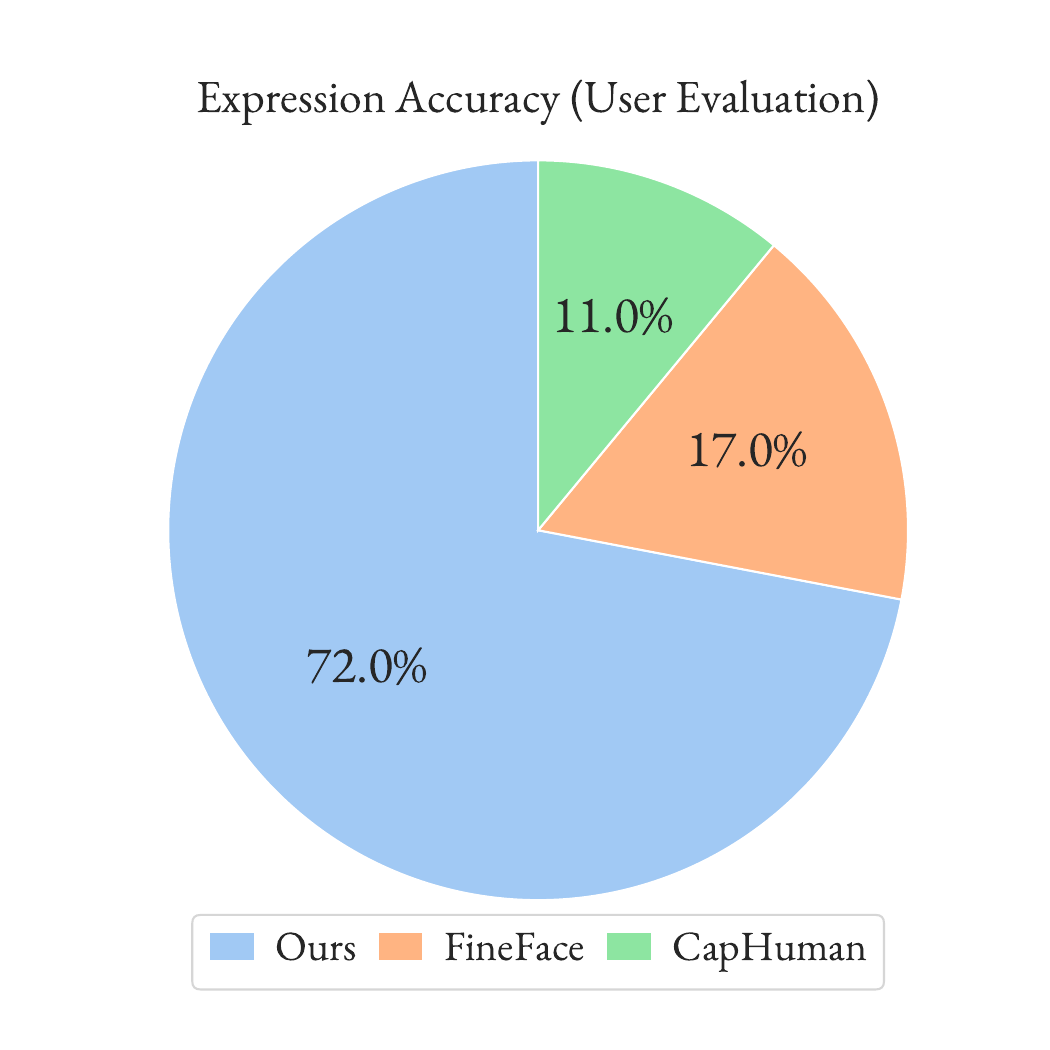}
      \caption{Users' preference on accuracy between generated/intended expressions.}
      \label{fig:user_study}
  \end{minipage}
\end{figure}

As shown in \cref{tab:quant}, our method significantly outperforms the baselines in terms of expression consistency across all metrics, while also achieving a lower FID score, indicating higher visual quality. This highlights the effectiveness of our precise, parametric expression representation in accurately transferring expressions against AU-based and render-based alternatives. Moreover, as illustrated in \cref{fig:id_sim}, our model consistently achieves higher identity similarity to the input subjects, owing to the strong Arc2Face prior and our careful integration of expression conditioning avoiding interference with identity. To further validate the superiority of our approach, we conducted a user study with 37 participants who were asked to choose the method whose result best resembles the desired expression across a randomly chosen set of 25 ID/expression pairs from our dataset. As reported in \cref{fig:user_study}, our method received 72\% of the votes, indicating its strong expression fidelity. For a visual comparison, we provide examples in \cref{fig:comp}. It is evident that other methods struggle to transfer varying target expressions to the input subjects, whereas our method performs reliably, even for subtle or uncommon facial expressions.

\begin{table}
    \begin{center}
    \setlength{\tabcolsep}{2.3pt}
    \scriptsize
    \begin{tabular}{lccccc} 
        \toprule
         & FID$\downarrow$ & Em.~Acc.~(\%)$\uparrow$ & AU-MSE$\downarrow$ & VA-MSE$\downarrow$ & 3D Exp-MSE$\downarrow$\\
         \midrule
         FineFace \cite{varanka2024fineface} & 0.367 & 35.85 & 0.038 & 0.164 & 1.250\\
         CapHuman \cite{liang2024caphuman} & 0.701 & 23.50 & 0.045 & 0.208 & 1.076\\
         \textbf{Ours} & \textbf{0.300} & \textbf{46.59} & \textbf{0.032} & \textbf{0.086} & \textbf{0.696}\\
        \bottomrule
    \end{tabular}
    \end{center}   
    \caption{\textbf{ID-conditioned generation:} Comparison of our method with ID-conditioned diffusion models for expression control \cite{varanka2024fineface, liang2024caphuman} on a dataset of 1K IDs. For each ID, we generate 5 expressive samples using target expressions shared across all methods. We report expression accuracy using metrics based on 3D reconstruction, emotion classification, Action Unit (AU), and Valence-Arousal (VA) predictions. Image quality is assessed using FID. Bold values indicate the best performance for each metric.}
    \label{tab:quant}
\end{table}

\subsection{Reference-Driven Expression Generation}
We further evaluate our method in a reference-driven setting, where the objective is to modify the facial expression of a given image while preserving the subject’s appearance and background. The evaluation setup mirrors that of the stochastic generation experiment, using the same metrics and test data. However, in this case, we activate the \textbf{Reference Adapter} and its associated LoRA layers for our method, allowing our model to condition not only on the ArcFace embedding but also directly on the source image itself. We compare against recent methods designed for this task: MagicFace \cite{wei2025magicface} which incorporates Action Units along with reference image features into the Stable Diffusion pipeline and Face-Adapter \cite{han2024faceadapter} which uses 3D landmarks instead, along with the source image. Finally, we include EmoStyle \cite{EmoStyle_2024_WACV}, a GAN-based approach which manipulates StyleGAN2 \cite{karras2020analyzing} latent codes using Valence-Arousal controls via a learned expression-guidance network.

\begin{figure}[h]
\centering
\includegraphics[width=1.0\linewidth, trim={5.0cm 0.8cm 4.0cm 1.0cm}, clip]{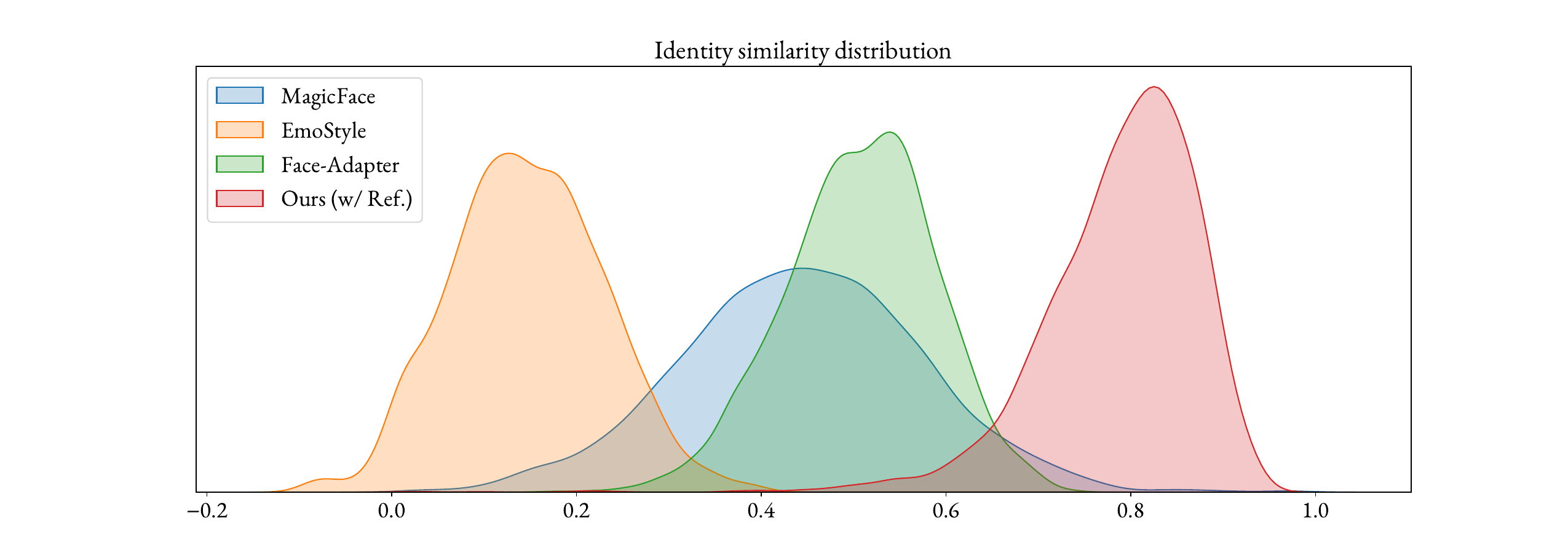}
\caption{Comparison of cosine similarity distributions between identity features of input and generated faces for the reference-driven setting.}
\label{fig:id_sim_ref}
\end{figure}

\begin{figure}[h]
\centering
\includegraphics[width=1.0\linewidth]{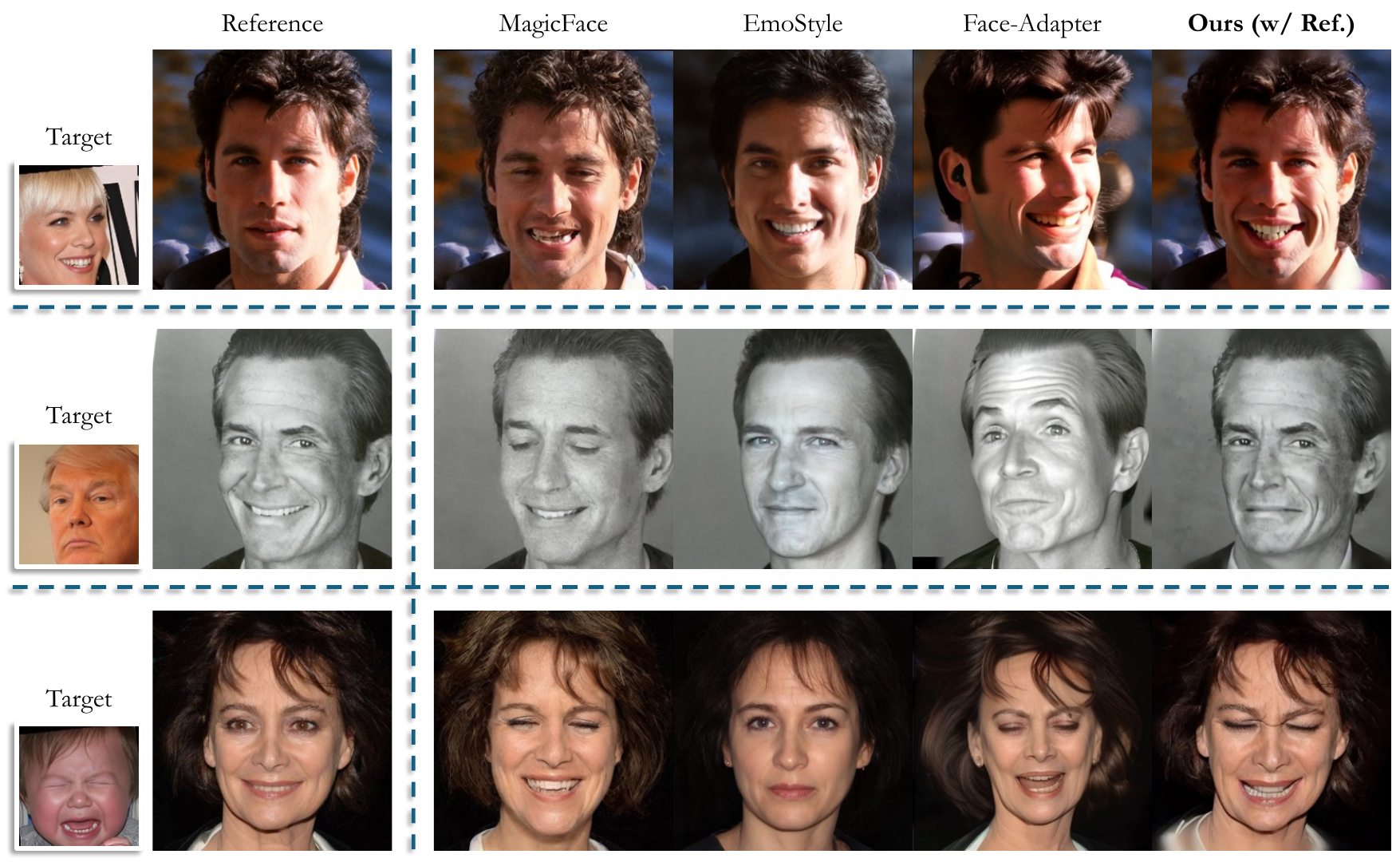}
\caption{Visual comparison with recent methods for reference-driven expression control~\cite{wei2025magicface, han2024faceadapter, azari2024emostyle}. Our method achieves more faithful expression transfer while better preserving both identity and visual consistency with the reference image.}
\label{fig:comp_ref}
\end{figure}

We present the results in \cref{tab:quant_ref} and \cref{fig:id_sim_ref} and provide visual comparisons in \cref{fig:comp_ref}. Our method again achieves superior results than the baselines in all aspects. While our primary design targets stochastic generation with expression and identity guidance, we show how our architecture can be seamlessly extended to reference-based editing via the proposed optional mechanism, outperforming models explicitly designed for this task.

\begin{table}[h]
    \begin{center}
    \setlength{\tabcolsep}{1.8pt}
    \scriptsize
    \begin{tabular}{lccccc} 
        \toprule
         & FID$\downarrow$ & Em.~Acc.~(\%)$\uparrow$ & AU-MSE$\downarrow$ & VA-MSE$\downarrow$ & 3D Exp-MSE$\downarrow$\\
         \midrule
         MagicFace \cite{wei2025magicface} & 0.138 & 34.08 & 0.044 & 0.166 & 1.595\\  
         EmoStyle \cite{azari2024emostyle} & 0.159 & 42.42 & 0.043 & 0.091 & 1.223\\ 
         Face-Adapter \cite{han2024faceadapter} & 0.166 & 43.85 & 0.040 & 0.093 & 0.986\\
         \textbf{Ours (w/ Ref.)} & \textbf{0.128} & \textbf{44.19} & \textbf{0.038} & \textbf{0.090} & \textbf{0.849}\\
        \bottomrule
    \end{tabular}
    \end{center}   
    \caption{\textbf{Reference-driven generation:} Comparison of our method - extended with the Reference Adapter - against approaches for expression editing in reference images \cite{wei2025magicface, EmoStyle_2024_WACV, han2024faceadapter}. The evaluation setup is identical to that in \cref{tab:quant}.  Bold values indicate the best performance for each metric.}
    \label{tab:quant_ref}
\end{table}

\subsection{Ablation Studies}
To further validate the effectiveness of our expression representation, we conduct ablation studies comparing our method against two alternative variants. First, we replace the FLAME blendshape parameters $\mathbf{s} \in \mathbb{R}^{55}$ with deep perceptual expression features. Specifically, we use EmoNet~\cite{toisoul2021estimation}, a widely used expression recognition model, to extract perceptual features $c \in \mathbb{R}^{256}$ from its final convolutional layers, just before classification. These features are then used as input to the \textbf{Expression Adapter}, instead of the original 3DMM-based parameters. Second, we evaluate a straightforward baseline that integrates ControlNet \cite{controlnet} with the Arc2Face model \cite{paraperas2024arc2face}. For this, we use the official pre-trained ControlNet variant provided by Arc2Face, which is conditioned on normal maps derived from the FLAME 3D model. We evaluate both baselines using the same expression consistency protocol as in the main experiments. \cref{tab:ablation} shows that both alternatives fall short in accurately encoding the intended expressions, leading to a noticeable performance drop. We attribute this to the highly discriminative nature of emotion recognition networks in the first case and ControlNet’s reliance on low-dimensional geometric cues, such as normal maps, in the second case, which provide only coarse surface information about facial structure. In contrast, the use of explicit and compact blendshape parameters, while simple in design, proves to be the most efficient approach for capturing and re-generating detailed expressions.

\begin{figure}[h]
\centering
\includegraphics[width=1.0\linewidth]{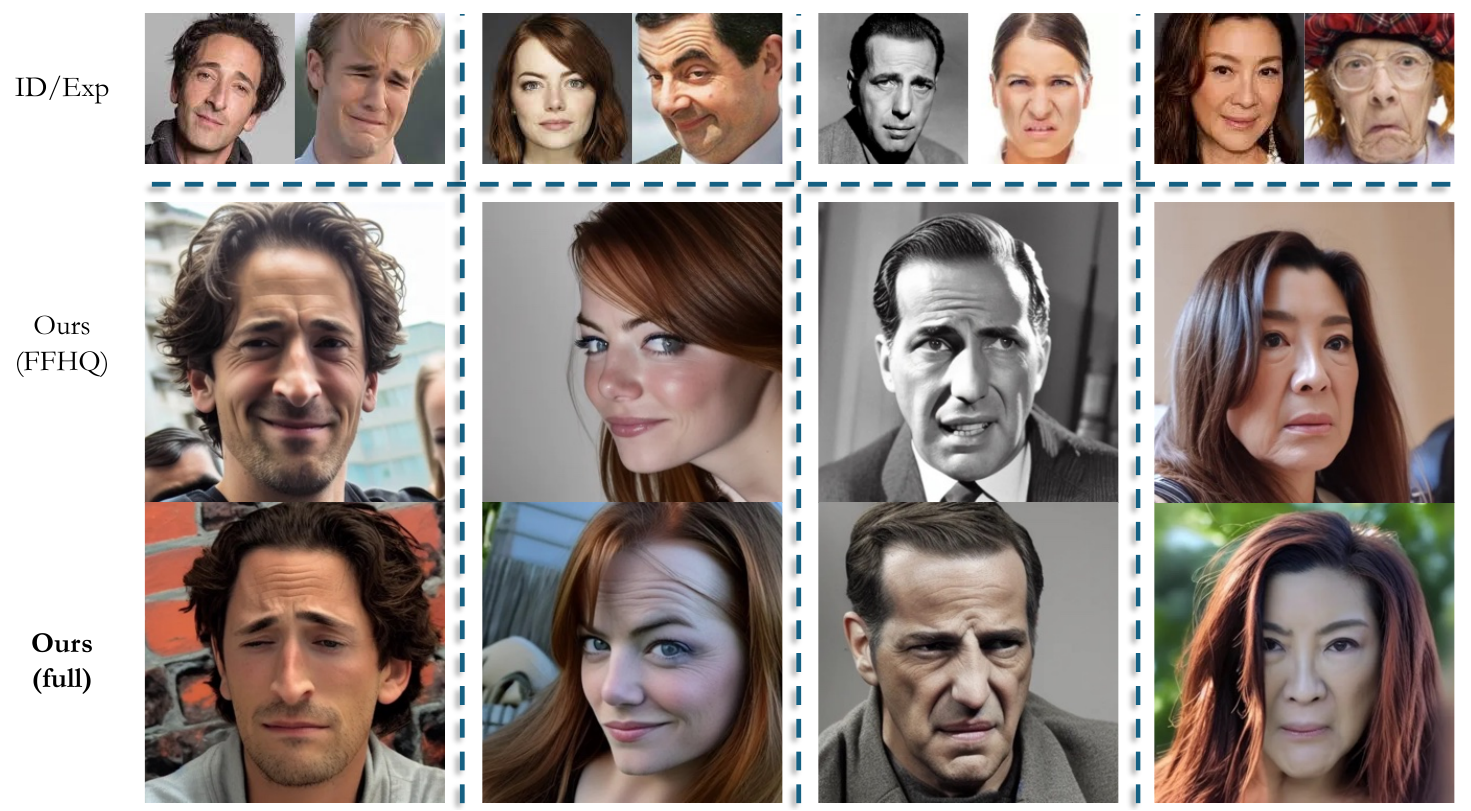}
\caption{Comparison of our Expression Adapter trained on the full dataset (AffectNet + FEED + FFHQ) vs.~trained only on FFHQ.}
\vspace{-0.3cm}
\label{fig:ablation}
\end{figure}

Furthermore, we examine the importance of expressive variability in the training data. As discussed earlier, in addition to FFHQ, we incorporate AffectNet~\cite{mollahosseini2017affectnet}, which is tailored for facial expression recognition (FER), and FEED~\cite{Drobyshev_2024_CVPR}, a dataset featuring exaggerated expressions performed by actors. These datasets are crucial for capturing a broad range of facial expressions and enabling the model to generalize beyond subtle or common emotions. To demonstrate this, we retrain our model using only the high-quality but expression-limited FFHQ dataset. \cref{fig:ablation} provides examples where this model fails to accurately reproduce more nuanced reference expressions that go beyond prototypical emotions such as happiness or anger, and instead reflect micro-expressions or facial morphs. The integration of AffectNet and FEED proves essential for robust expression synthesis across the full spectrum of human affect.

\begin{table}[h]
    \begin{center}
    \setlength{\tabcolsep}{2.5pt}
    \scriptsize
    \begin{tabular}{lccccc} 
        \toprule
         & Em.~Acc.~(\%)$\uparrow$ & AU-MSE$\downarrow$ & VA-MSE$\downarrow$ & 3D Exp-MSE$\downarrow$\\
         \midrule
         Arc2Face + ControlNet & 31.92 & 0.038 & 0.154 & 1.167\\
         Ours w/ EmoNet & 44.28 & 0.037 & 0.105 & 1.192\\
         \textbf{Ours} & \textbf{46.59} & \textbf{0.032} & \textbf{0.086} & \textbf{0.696}\\
        \bottomrule
    \end{tabular}
    \end{center}
    \vspace{-0.3cm}
    \caption{Ablation study comparing the proposed approach with two alternative variants: (1) integrating a ControlNet into Arc2Face, using normal maps for control instead of expression blendshapes; and (2) using the proposed Expression Adapter, but conditioned on deep features from a Facial Expression Recognition (FER) network rather than explicit 3DMM parameters.}
    \label{tab:ablation}
\end{table}

\section{Limitations}
\label{sec:limitations}

\begin{figure}[h]
\centering
\includegraphics[width=1.0\linewidth]{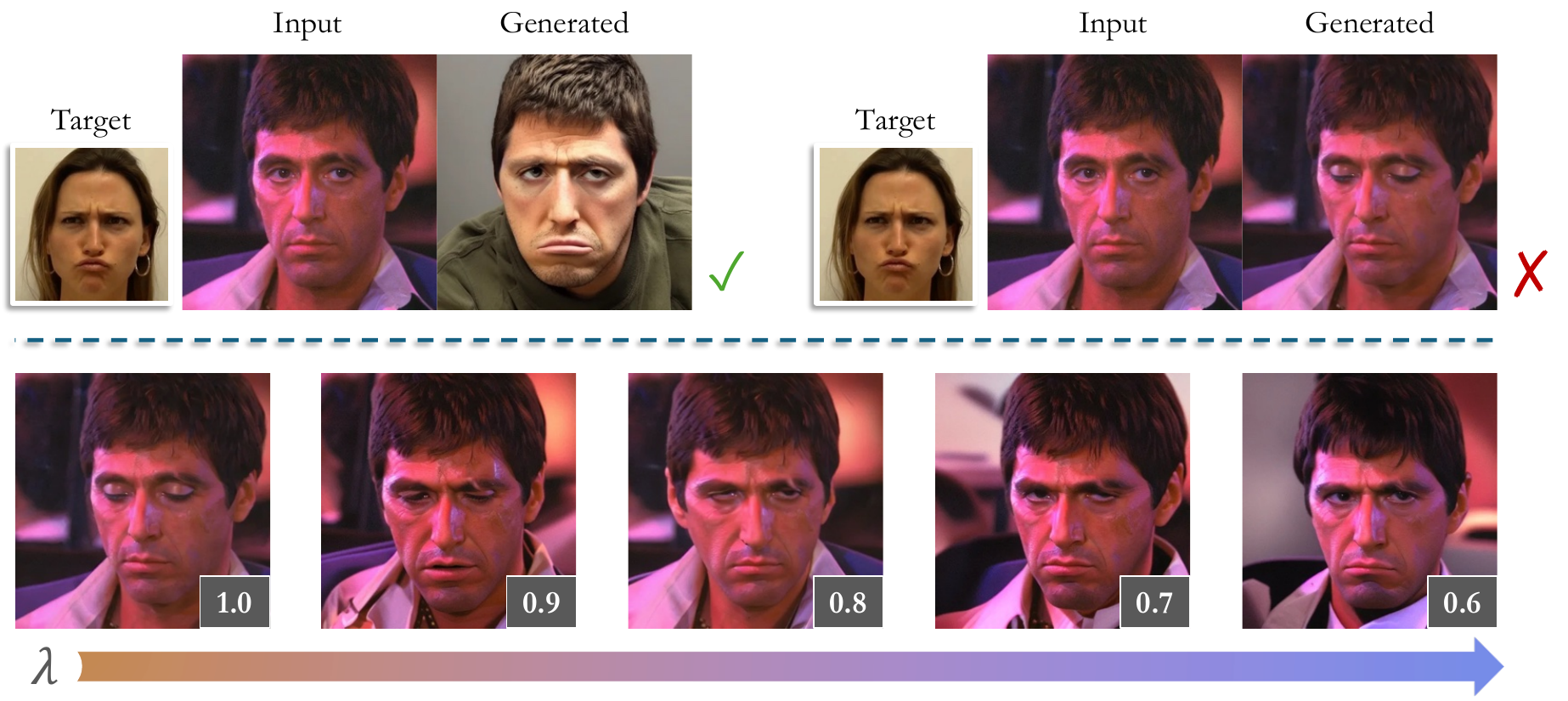}
\caption{\textbf{Top:} Example of failure case where the base model successfully reproduces the target expression, but the reference-driven variant struggles due to over-reliance on the source image. \textbf{Bottom:} Reducing the LoRA scaling factor $\lambda$ of the Reference Adapter alleviates this by trading off pose and background consistency for improved expression fidelity.}
\label{fig:failure_cases}
\end{figure}

While our approach effectively generates precise and detailed expressions, some limitations remain. First, the adopted parametric representation lacks semantic interpretability. As a result, expression manipulation typically relies on extracting parameters from reference images, which may be limiting for certain applications. Second, the accuracy of expression reproduction is inherently dependent on the quality of the 3D reconstruction method used to extract blendshape parameters~\cite{smirk_2024_CVPR}, which although state-of-the-art might still be imperfect and introduce occasional errors. Finally, expression editing with the \textbf{Reference Adapter} can sometimes be inconsistent. As shown in \cref{fig:failure_cases} (top), there are instances where the base model accurately transfers the target expression, but the reference-driven variant fails to do so. This is due to the \textbf{Reference Adapter} often tending to ``copy-paste'' the reference image, despite our use of cross-paired video data for training. This behavior can be mitigated at test time by adjusting the scaling factor $\lambda$ of the LoRA layers, which modulates the influence of the reference image. As shown in \cref{fig:failure_cases} (bottom), reducing $\lambda$ improves expression fidelity at the cost of slight deviations in background or pose from the reference image.
\\
\noindent \textbf{Note on social impact.} We acknowledge the ethical implications of our work, as technologies for controllable face generation may be misused to produce deceptive or unoriginal facial imagery, particularly when capable of altering expressions. While our goal is to advance research in positive domains such as accessibility and creative storytelling, we recognize these risks and emphasize the importance of ethical standards and investment in countermeasures like synthetic content detection.

\vspace{-0.3cm}
\section{Conclusion}
\label{sec:conclusion}

In this work, we extend a foundation face model, Arc2Face, to enable explicit and accurate expression control in the context of ID-consistent face generation. Our \textbf{Expression Adapter} maps expression parameters from a 3DMM into the latent space of CLIP embeddings, aligning with Stable Diffusion’s conditioning interface, while its dual attention mechanism enables disentangled expression control without compromising identity fidelity. Given blendshape parameters extracted from a target image via 3D reconstruction, our model faithfully reproduces the same expression to any source identity, with superior accuracy and visual quality than prior methods. We place particular emphasis on curating expression-rich training data to provide a robust model that goes beyond simple emotions, enabling complex expressive transitions that lie between canonical emotional states. We further enable image-based expression editing through a \textbf{Reference Adapter} that conditions the generation on multi-scale features from the reference image via an adapted self-attention mechanism. Our open-source model can provide a versatile tool for controllable synthetic data generation, benefiting downstream tasks such as Facial Expression Recognition (FER), while bridging the gap between general-purpose image generation and human-centric synthesis for AI-driven storytelling.
\\
{\sloppy
\textbf{Acknowledgements.} S. Zafeiriou and part of the research was funded by the EPSRC Project GNOMON (EP/X011364/1) and Turing AI Fellowship (EP/Z534699/1).
\par
}

{
    \small
    \bibliographystyle{ieeenat_fullname}
    \bibliography{main}
}

\clearpage
\onecolumn
\appendix

\begin{center}
\vspace*{8pt}
\textbf{\Large ID-Consistent, Precise Expression Generation with Blendshape-Guided Diffusion}

\vspace{3pt}
\textbf{\Large (Supplementary Material)}
\vspace{20pt}
\end{center}

\section{Implementation Details}
\label{sec:implementation}

Our method builds on Arc2Face~\cite{paraperas2024arc2face}, which uses a fine-tuned UNet and ID encoder derived from \texttt{stable-diffusion-v1-5}. For our \textbf{Expression Adapter}, we employ a two-layer MLP, as well as separate identical key/value matrices into the UNet's cross-attention layers. Training is performed with AdamW~\cite{loshchilov2017decoupled} using a learning rate of 1e-4, a batch size of 8 per GPU, across 8 NVIDIA A100 GPUs for 300K iterations. For the \textbf{Reference Adapter}, we use a copy of Arc2Face’s UNet as the reference network and augment the original UNet with LoRA matrices of rank 128. The LoRA weights are trained using cross-paired frames from video datasets, as described in the main paper. We optimize them with a learning rate of 1e-5, using the same hardware configuration and a batch size of 8, for 15K iterations. For inference, we adopt DPM-Solver~\cite{lu2022dpm, lu2022dpm++} with 25 denoising steps and a classifier-free guidance scale of 3. For the LoRA weights in the \textbf{Reference Adapter}, we found a scale factor of 0.8 to provide a good balance between visual consistency with the input image and alignment with the target expression. Finally, for the baselines in our comparisons, we replace any additional text conditioning (if used by the method) with the prompt “photo of a person” to ensure a fair comparison that emphasizes identity and expression control.

\section{Additional Results}

\begin{figure*}[h]
\centering
\includegraphics[width=0.88\textwidth]{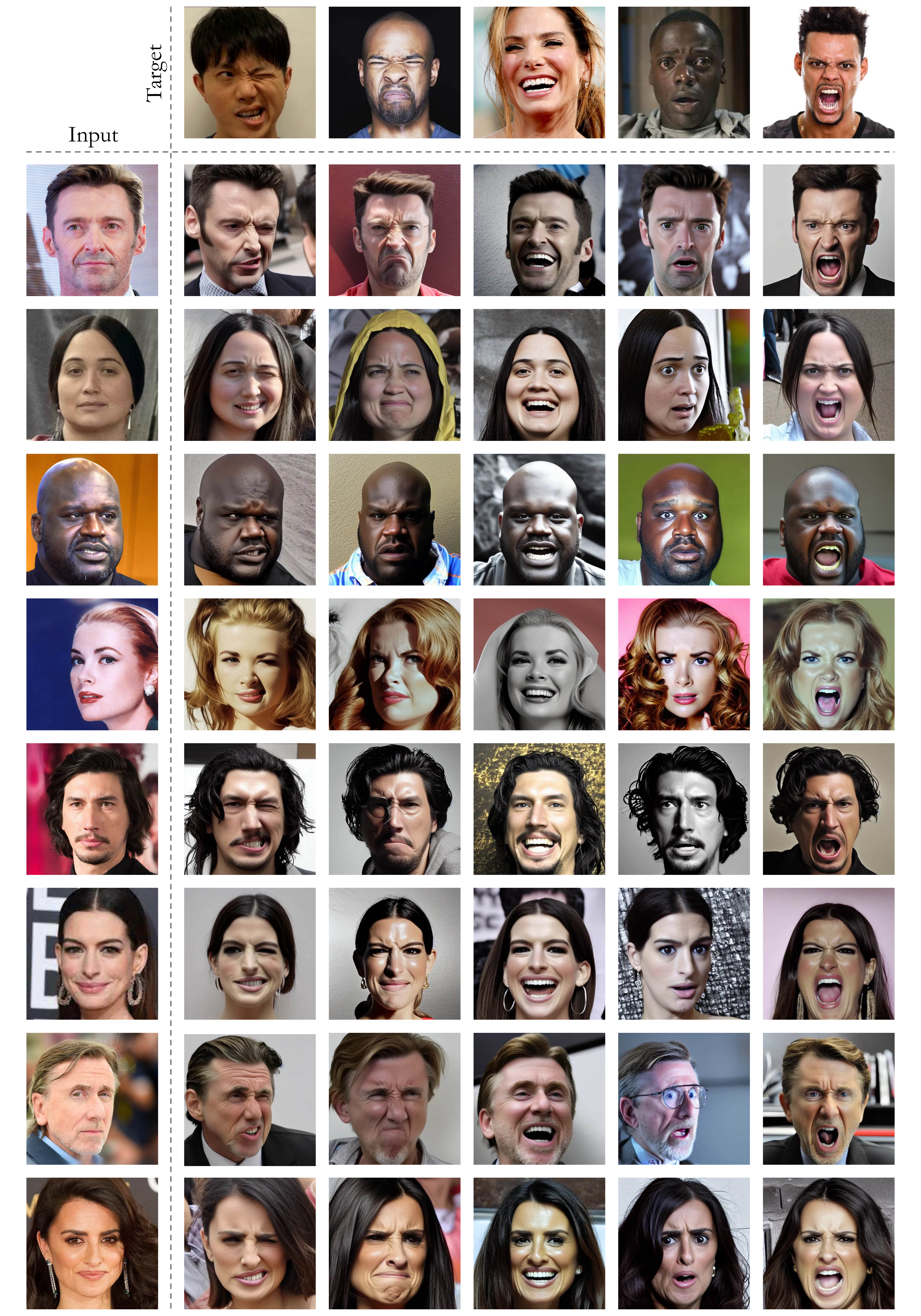}
\caption{Samples generated by our model, conditioned on the input identity and the corresponding target expression.}
\label{fig:samples1}
\end{figure*}

\begin{figure*}[h]
\centering
\includegraphics[width=0.88\textwidth]{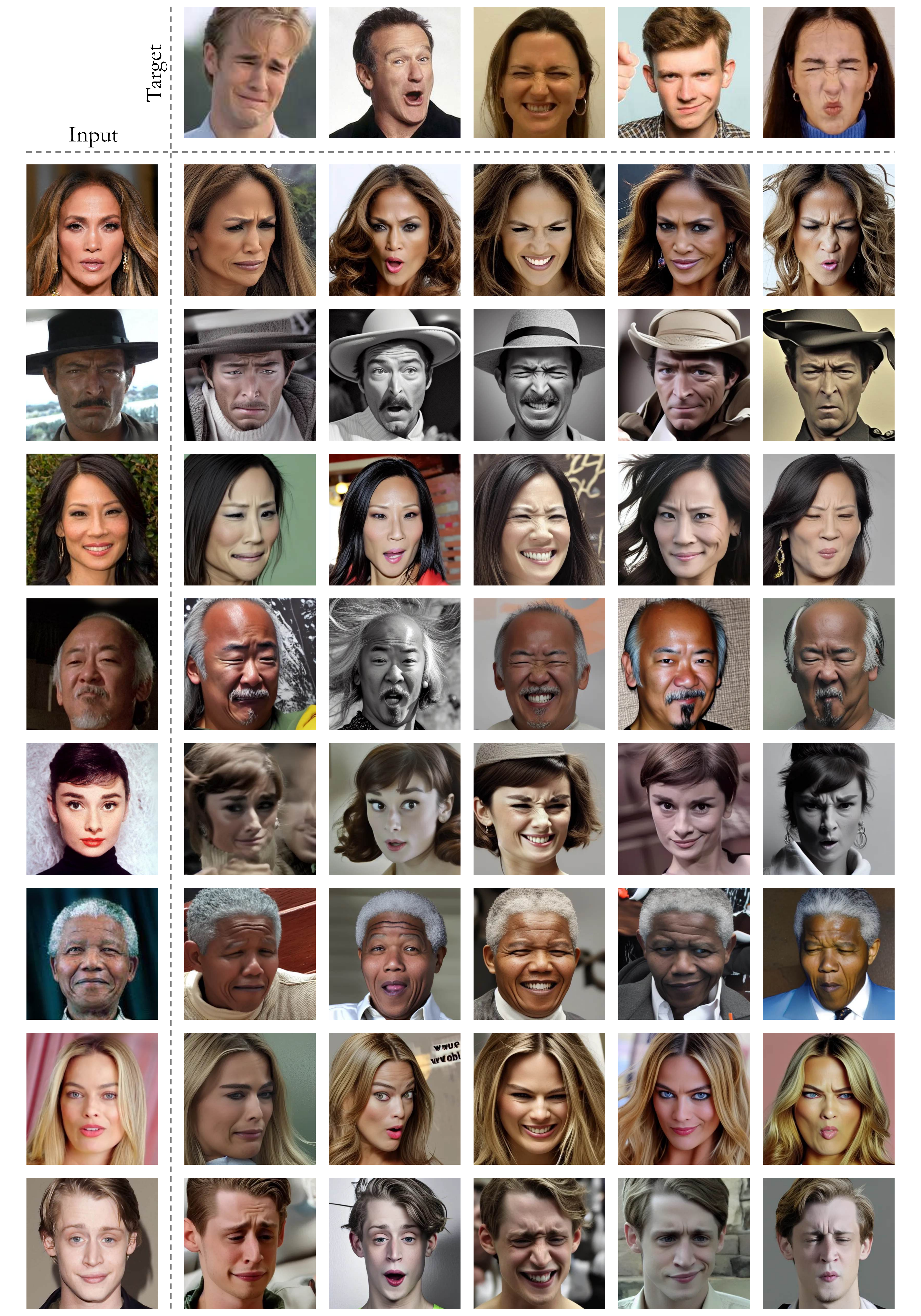}
\caption{Samples generated by our model, conditioned on the input identity and the corresponding target expression (cont.).}
\label{fig:samples2}
\end{figure*}

\begin{figure*}[h]
\centering
\includegraphics[width=0.88\textwidth]{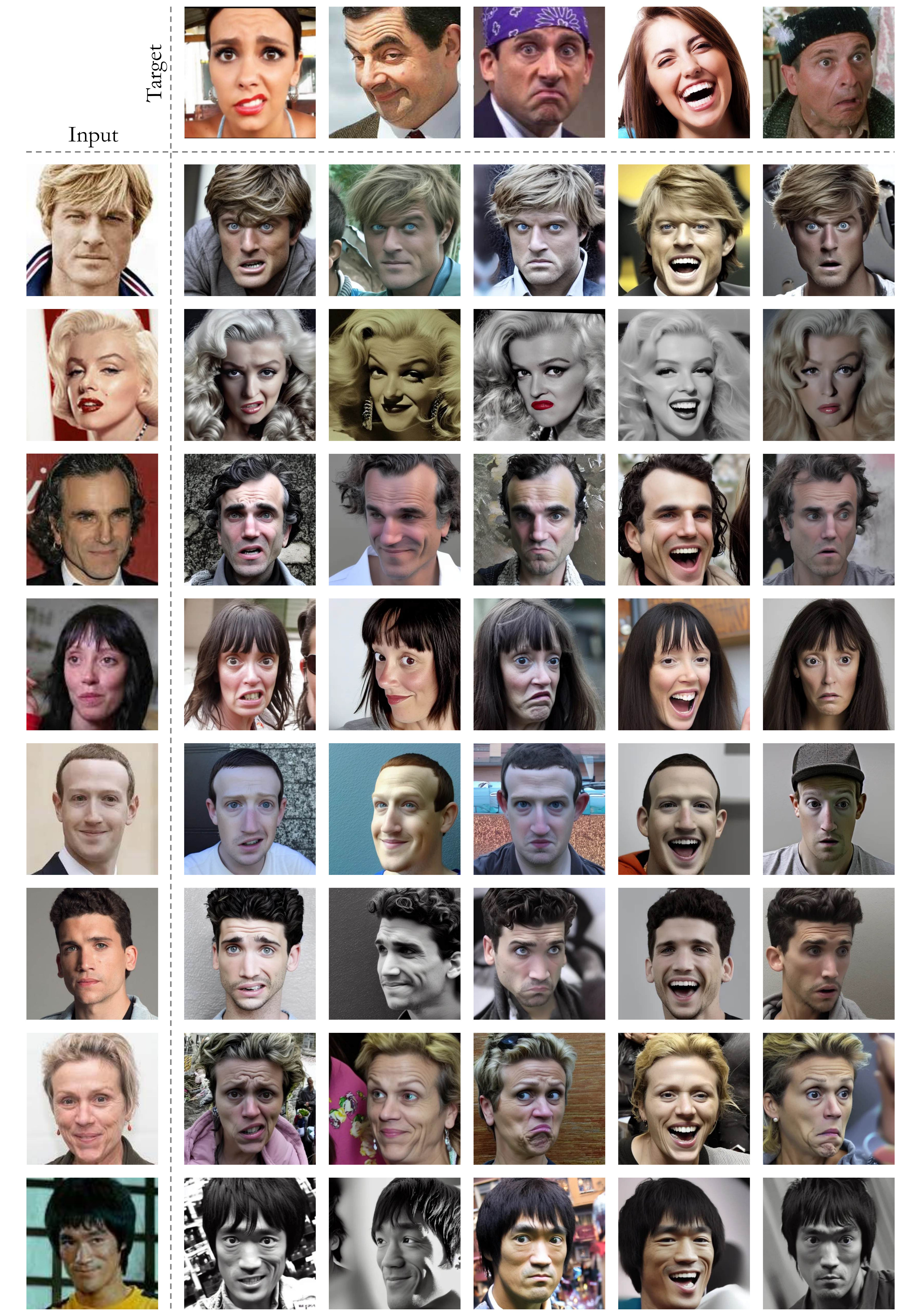}
\caption{Samples generated by our model, conditioned on the input identity and the corresponding target expression (cont.).}
\label{fig:samples3}
\end{figure*}

Below, we provide additional visual comparisons in \cref{fig:comp_supp}, along with further generations from our \textbf{Expression Adapter} in \cref{fig:samples1,fig:samples2,fig:samples3}.

\begin{figure*}[h]
\centering
\includegraphics[width=0.87\textwidth]{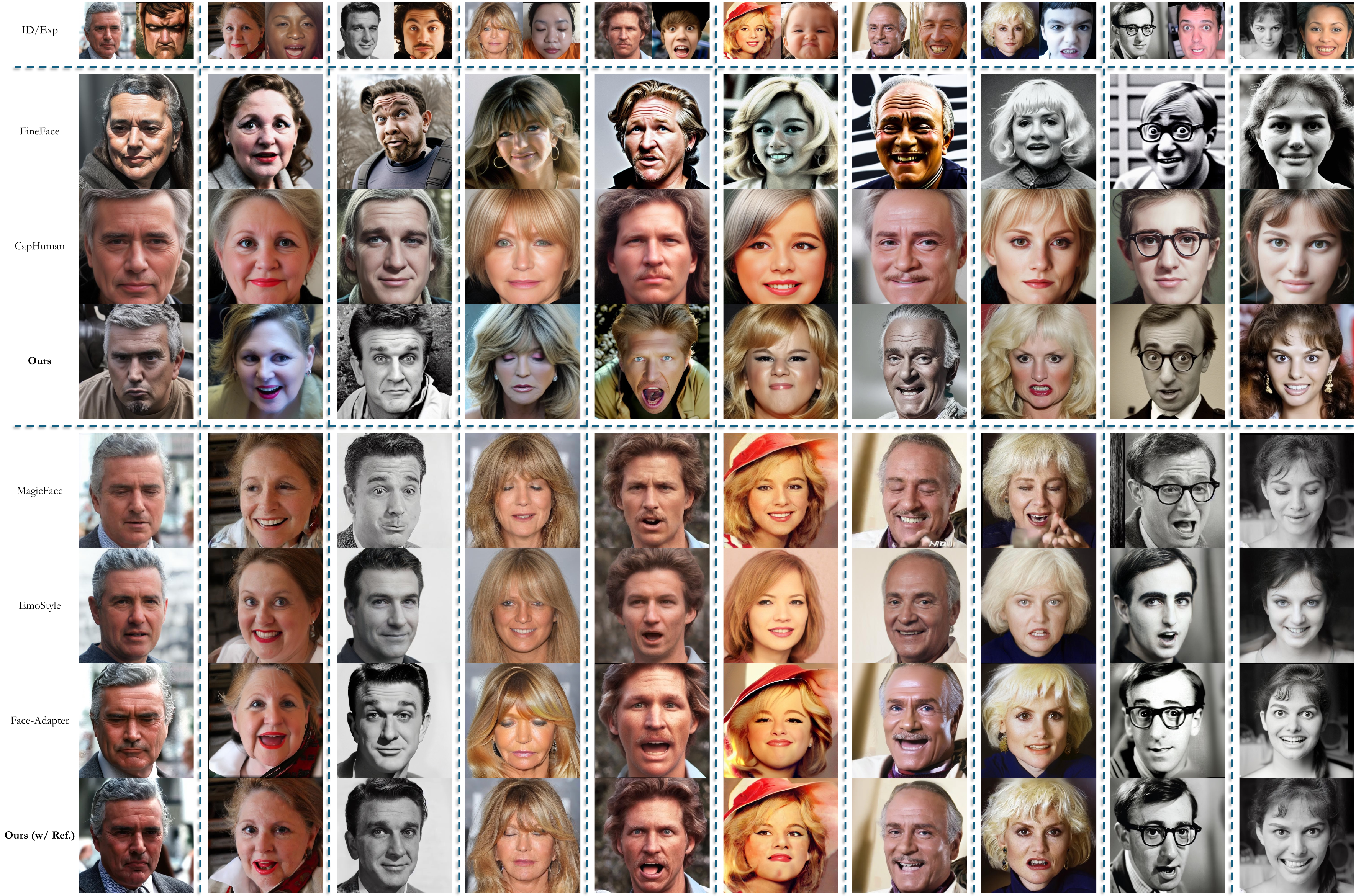}
\caption{Visual comparison between our method and competing expression-conditioned models. \textbf{Top:} ID-driven results compared to \cite{varanka2024fineface, liang2024caphuman}. \textbf{Bottom:} Reference-driven generation compared to \cite{wei2025magicface, azari2024emostyle, han2024faceadapter}. For the latter setting, our method is conditioned on both identity features and the reference image via the Reference Adapter.}
\label{fig:comp_supp}
\end{figure*}

\end{document}